\documentclass{llncs}
\usepackage{makeidx}  

\usepackage{graphicx}
\usepackage{bm,amsmath,amsfonts,amssymb,dsfont} 
\usepackage{hyperref} 

\usepackage{subcaption}

\graphicspath{{figures/}}

\begin{document}
\frontmatter          
\pagestyle{headings}  
\addtocmark{Grasp Pose Generation} 
\mainmatter              
\title{Automatic Grasp Pose Generation for Parallel Jaw Grippers}
\titlerunning{Grasp Pose Generation}  
%
\author{Kilian Kleeberger\inst{1} \and Florian Roth\inst{1} \and Richard Bormann\inst{1} \and Marco F. Huber\inst{1,2}}
\authorrunning{Kilian Kleeberger et al.} 
%
\tocauthor{Kilian Kleeberger, Florian Roth, Richard Bormann, Marco F. Huber}
\institute{Fraunhofer Institute for Manufacturing Engineering and Automation IPA, Nobelstra{\ss}e~12, 70569 Stuttgart, Germany,\\
\email{kilian.kleeberger@ipa.fraunhofer.de},\\ WWW home page:
\texttt{https://www.bin-picking.ai/}
\and
Institute of Industrial Manufacturing and Management IFF, University of Stuttgart, Allmandring~35, 70569 Stuttgart, Germany,\\
\email{marco.huber@ieee.org}}

\maketitle              

\begin{abstract}
This paper presents a novel approach for the automatic offline grasp pose synthesis on known rigid objects for parallel jaw grippers. We use several criteria such as gripper stroke, surface friction, and a collision check to determine suitable 6D grasp poses on an object. In contrast to most available approaches, we neither aim for the best grasp pose nor for as many grasp poses as possible, but for a highly diverse set of grasps distributed all along the object. In order to accomplish this objective, we employ a clustering algorithm to the sampled set of grasps. This allows to simultaneously reduce the set of grasp pose candidates and maintain a high variance in terms of position and orientation between the individual grasps. We demonstrate that the grasps generated by our method can be successfully used in real-world robotic grasping applications.



\end{abstract}

\section{Introduction}





Robots are extensively used in modern production and significantly increase the efficiency in various areas like assembly, sorting, and handling~\cite{Litvak.2019}. However, from a production standpoint, especially part handling does not contribute to the added value of products~\cite{Wolf.2005}. A considerable increase in handling efficiency is obstructed by several factors. The limited environmental awareness of most in-the-field robots requires the manufacturing products to be delivered in strictly defined poses. Furthermore, a quick change of the manufactured product is often associated with extensive reprogramming of the robotic system~\cite{MolzowVoit.2016,Litvak.2019,Muller.2019,VogelHeuser.2017}. Numerous work places today still operate manually and their poor ergonomics or basic
and monotonous tasks still reveal existing obstacles for a fully automated factory. Even a seemingly simple task such as grasping and retrieving an object from a bin with chaotically stored objects remains a major problem in robotics~\cite{Wolf.2005,Buchholz.2016,Kleeberger_PQNet.2020,Kleeberger_ReviewArticle.2020}. To enable an increased efficiency in handling tasks, the so-called bin-picking problem needs to be solved. Bin-picking consists of several sub-challenges. Amongst others, the pose of the objects needs to be estimated and suitable grasps have to be generated successively. These grasps have to be kinematically feasible, without collision, and able to lift the object~\cite{Pas.2015,Shao.2018,ZapataImpata.2019}. The automated generation of grasp poses (or gripping points) presents a suitable approach to reduce manual configuration and increase efficiency in bin-picking tasks~\cite{Kundu.2018}. Accordingly, a decreased manual configuration effort enables a greater scalability and economic efficiency of bin-picking applications.

In this paper, we present an approach for the automatic grasp pose generation for known rigid objects. A point cloud is used as input, consisting of points sampled from the surface of the object and the corresponding normals. Any two-point combination of an input point cloud is checked with respect to various metrics. If all tests are successful, we can assume to have obtained a possible grasp pose candidate. As the resulting amount of possible grasps through this procedure can be overwhelming, we introduce a clustering algorithm to enable a reduction of the grasp poses while maintaining a high variance.

In summary, the main contributions of this work are:
\begin{itemize}
\item Novel approach for automatic grasp pose generation for parallel jaw grippers
\item Grasp clustering for highly diverse grasps on the object model (no local concentrations)
\item Provide grasp poses for the objects from various public datasets
\item Experimental evaluation of the generated grasp poses in a real-world robot cell
\end{itemize}

The rest of this paper is structured as follows. First, we evaluate various approaches of the literature regarding grasp synthesis in Section~\ref{sec:related}. Section~\ref{sec:auto_generation} introduces our approach for grasp generation originating from~\cite{MA_FlorianRoth}. Using several criteria, we show the process of sampling a large quantity of grasp poses from a part geometry. To facilitate a reduced set of grasps while maintaining a high variance between individual grasp poses, we employ a clustering algorithm in the pose space. The resulting grasps are evaluated in Section~\ref{sec:evaluation}, both qualitatively and on a real-world robot cell.
Finally, we discuss strengths and limitations of our approach in Section~\ref{sec:discussion} and draw conclusions in Section~\ref{sec:conclusion}.

\section{Related Work}
\label{sec:related}
\subsection{Basic Contact Principles}




\begin{figure}[tb]
    \centering
    \begin{subfigure}{0.3\textwidth}
        \centering
        \includegraphics[width=0.5\textwidth]{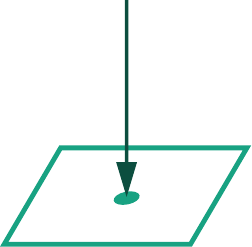}
        \subcaption{}
    \end{subfigure}
    \hfill
    \begin{subfigure}{0.3\textwidth}
        \centering
        \includegraphics[width=0.5\textwidth]{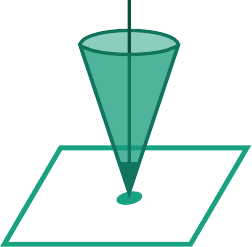}
        \subcaption{}
    \end{subfigure}
    \hfill
    \begin{subfigure}{0.3\textwidth}
        \centering
        \includegraphics[width=0.5\textwidth]{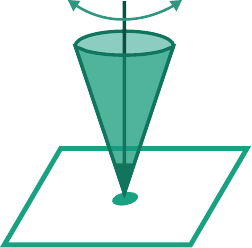}
        \subcaption{}
    \end{subfigure}
    \caption{Contact models: (a) rigid body frictionless, (b) rigid body with friction, (c) soft body with friction~\cite{Borst.2003}.}
    \label{fig:contact_model}
\end{figure}

Various methods for generating grasp poses have been proposed. Yet, the basics are often alike. Reduced to a single point, three kinds of contact can be differentiated, which are illustrated in Fig.~\ref{fig:contact_model}. Fig.~\ref{fig:contact_model}~(a) shows the resulting contact of a rigid body to a surface without friction. Thus, only forces in the direction of the surface normal can be absorbed. Fig.~\ref{fig:contact_model}~(b) shows rigid body contact with friction. All forces within the displayed friction cone can be absorbed. The friction cone is defined via the friction $\mu$ to $ \alpha=\arctan(\mu) $. For the case of a soft body, additional torsion forces can be absorbed as displayed in Fig.~\ref{fig:contact_model}~(c). Though, in the more realistic scenario of more than one contact point, the last example can be disregarded~\cite{Borst.2003,Siciliano.2016}.

\subsection{Grasp Synthesis}
The bin-picking problem has been researched for several decades. To enable robotic systems for grasping and placing objects of various geometries, several sub-problems have to be solved. Among others, the research area consists of scene analysis, object detection, pose estimation, grasp planning, and path planning~\cite{Buchholz.2016,Siciliano.2016,Kleeberger_ReviewArticle.2020}. In contrast to image processing and path planning, the synthesis of grasp poses seems to be of proprietary nature more often~\cite{Correll.2016}. Nevertheless, finding suitable grasp poses for the object of choice is an important step in bin-picking. Various methods for the generation of grasps have been proposed~\cite{Sahbani.2012,Bohg.2014}. 

Generally, these methods can be divided into analytical and data-driven methods. While the former often use simple metrics, such as Coulomb friction for grasp generation, the latter mostly execute a huge amount of grasps in a simulated environment in search of successful ones. Under close examination, even approaches aiming at the creation of an artificial neural network for grasp synthesis often use analytical or randomized grasp synthesis methods to generate initial training data~\cite{Johns.2016,Mahler.2016,Mousavian.2019,Bohg.2011,Bohg.2014,Brook.2011}. Aside from analytical and randomized methods for grasp generation, hand-labeled data and teach-in methods are popular~\cite{Bohg.2014}. In all cases, these methods can be further distinguished in whether they aim at grasping known or unknown objects. But even when analysing machine learning based approaches for the grasping of unknown objects, initial learning data is often required~\cite{Mahler.2016,Mousavian.2019}.

While automatic grasp pose synthesis approaches increasingly gain popularity, many established bin-picking applications still utilize a manual grasp pose definition~\cite{FPS_GPC_1,FPS_GPC_2}. This usually requires expert knowledge of the object and application, leading to a work-intensive reconfiguration process~\cite{Spenrath.2017,Buchholz.2013,Spenrath.2013,Chen.2011,Schraft.2003}. Some of these approaches enable the user to add degrees of freedom around defined grasps to further increase the total number of grasp poses and simplify the teach-in process~\cite{Spenrath.2017,Buchholz.2013}.

\begin{center}
\begin{table}
    \centering
    \begin{tabular}{|l|c|c|c|}
        \hline
        \textbf{Approach} & \textbf{Geometry} & \textbf{Definition} & \textbf{Freedom} \\ [0.5ex]
        \hline\hline
        Mahler et al.~\cite{Mahler.2015} & complex & algorithmic & top-down  \\
        \hline
        Spenrath et al.~\cite{Spenrath.2013,FPS_GPC_1,FPS_GPC_2} & complex  & manual & 6D \\
        \hline
        Gatrell~\cite{Gatrell.1989} & simple & algorithmic & 6D  \\
        \hline
        Smith et al.~\cite{Smith.1999} & simple & algorithmic & top-down  \\
        \hline
        Buchholz et al.~\cite{Buchholz.2013} & complex & manual & 6D  \\
        \hline
        Mousavian et al.~\cite{Mousavian.2019} & complex & random & 6D  \\
        \hline
        Brook et al.~\cite{Brook.2011} & complex & random & 6D  \\
        \hline
        Bohg et al.~\cite{Bohg.2011} & complex & random & 6D  \\
        \hline
        Depierre et al.~\cite{Depierre.2018} & complex & random & top-down \\
        \hline
        Johns et al.~\cite{Johns.2016} & complex & random & top-down \\
        \hline
        Saxena et al.~\cite{Saxena.2008} & complex & manual & 6D \\
        \hline
        ours & complex & algorithmic & 6D \\
        \hline
    \end{tabular}
    \caption{Grasp synthesis approaches compared in terms of part geometry (complex or simple), grasp definition (manual, algorithmic, or random), and the degrees of freedom of the grasp (top-down or 6D).}
    \label{tab:approaches}
\end{table}
\end{center}


Especially some of the earlier approaches for the generation of grasps only use simple part geometries~\cite{Borst.2003,Ikeuchi.1983,Gatrell.1989}. One of the first was introduced by Gatrell~\cite{Gatrell.1989} in 1989. Using an extension of Gaussian images~\cite{Horn.1984}, Gatrell determines opposite surfaces on objects to efficiently search for suitable grasps. By neglecting grasps on concave surfaces and performing a collision check with a model of the used gripper, Gatrell further refines the amount of applicable grasps. The resulting grasps are then checked for rotational stability and sorted accordingly. Borst et al.~\cite{Borst.2003} create randomized grasp candidates consecutively reviewing them by different metrics, such as the grasp wrench space~\cite{Borst.2003} and the convex cone described by Hirai~\cite{Hirai.1991}. They highlight the efficiency of their approach for simple geometries.

Another form of simplification is the spatial reduction of synthesized grasp poses. This can have multiple reasons. Neglecting two of the rotational degrees of freedom, Smith et al.~\cite{Smith.1999} generate grasps for a SCARA robot kinematic, which is only capable of executing 4D grasps. In their approach, they use horizontal cross-sections of the object to be picked. In a multi-stage procedure, grasps are then generated along the edges of the created 2D polygon and different criteria such as friction and accessibility are checked. The resulting grasps are sorted by means of a metric consisting of friction and applicable torque. In a more recent approach, Mahler et al.~\cite{Mahler.2016} use an extension~\cite{Mahler.2015} of the grasp planning proposed by Smith et al.~\cite{Smith.1999} to generate grasps for known objects. Using similar metrics as Smith et al., they incorporate uncertainty in the underlying 3D camera images of a single viewpoint. With only a partial representation of the unknown geometry of the part, the grasps are therefore reduced to top-down grasps. Consecutively, they train an artificial neural network with the generated grasps. Similarly, Johns et al.~\cite{Johns.2016} use a fixed grasping height and define the longitudinal axis of the gripper strictly orthogonal to the floor. This results in 3D grasps with two translational and one rotational dimension, the latter around the gripper's longitudinal axis.

Table~\ref{tab:approaches} illustrates a comprehensive overview of different approaches and their underlying grasp synthesis methodologies. Online approaches using an artificial neural network for unknown objects often train these networks using labeled datasets. Thus, for these machine learning approaches only the offline training phase is considered.

Hence, most approaches use an offline grasp synthesis, either for an initial training data set or for the final execution. Beside an automatic generation of grasps and high dimensionality of the gripper pose, a high variance of the generated grasp poses is seen as beneficial~\cite{Mousavian.2019}. For most industrial applications, it can further be expected that the part geometry is available beforehand~\cite{Buchholz.2016,Correll.2016}. 

\section{Automatic Grasp Pose Generation}
\label{sec:auto_generation}

\begin{figure}[tb]
    \centering
    \begin{subfigure}[c]{0.4\textwidth}
        \centering
        \includegraphics[width=0.8\textwidth]{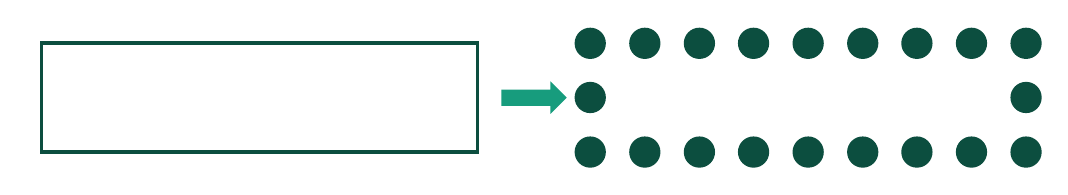}
        \subcaption{}
    \end{subfigure}
    \hfill
    \begin{subfigure}[c]{0.4\textwidth}
        \centering
        \includegraphics[width=0.8\textwidth]{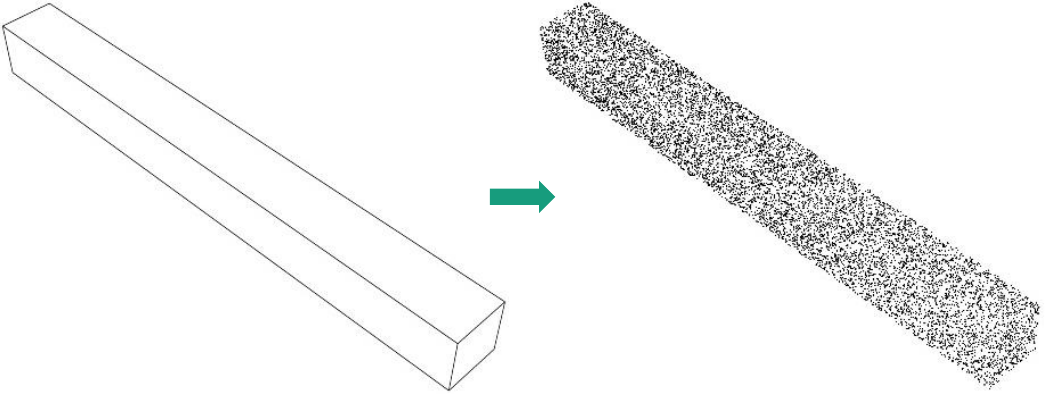}
        \subcaption{}
    \end{subfigure}
    \caption{Point sampling procedure shown (a) in principle and (b) on a real part.}
    \label{fig:point_sampling}
\end{figure}

We propose an offline approach for the automatic generation of grasps for a given object. In industrial use cases, we can either expect to obtain the parts' CAD data or a scan of the part a priori~\cite{Bormann.2019,Kasper.2012}. Therefore, it seems reasonable to use this knowledge to generate part-specific grasps. To create these, a set of criteria needs to be fulfilled. These criteria are defined to be necessary for a suitable grasp candidate. While most CAD file formats allow for a continuous description of the parts' surfaces, many applications use their own, proprietary file format. Converting them to open formats often inherits loss of geometric information either way. In addition, discretized geometries are easier to process. Therefore, in the following work, point clouds are used to represent the geometry of the parts as depicted in Fig.~\ref{fig:point_sampling} for both an artificial object and a real CAD model. These point clouds can be generated from CAD files. Furthermore, through the use of point cloud data, the straightforward use of part scans is possible. To enable a greater diversity of grasp poses, in a last step a clustering algorithm is used to reduce the resulting number of grasps. 

\begin{figure}[tb]
    \centering
    \begin{subfigure}{0.24\textwidth}
        \centering
        \includegraphics[height=1.25cm]{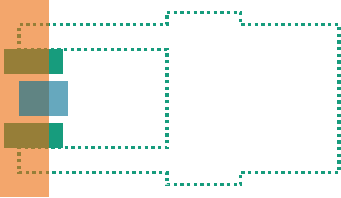}
        \subcaption{}
    \end{subfigure}
    \hfill
    \begin{subfigure}{0.24\textwidth}
        \centering
        \includegraphics[height=1.25cm]{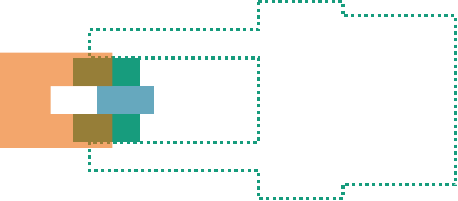}
        \subcaption{}
    \end{subfigure}
    \hfill
    \begin{subfigure}{0.24\textwidth}
        \centering
        \includegraphics[height=1.25cm]{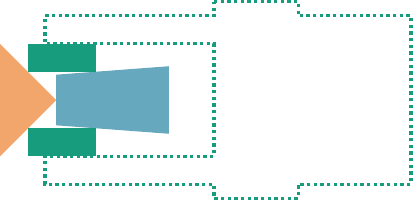}
        \subcaption{}
    \end{subfigure}
    \hfill
    \begin{subfigure}{0.24\textwidth}
        \centering
        \includegraphics[height=1.25cm]{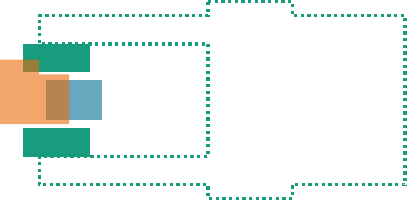}
        \subcaption{}
    \end{subfigure}
    \caption{Criteria that need to be fulfilled to consider a point combination as a valid grasp: (a) gripper stroke, (b) surface orientation, (c) surface friction, (d) collision. The gripper is indicated in green. Valid grasps are sketched in blue and infeasible ones in orange.}
    \label{fig:criteria}
\end{figure}

For each two-point combination, four criteria need to be fulfilled to form a valid grasp candidate. Quite trivially, for a successful grasp the part has to fit between the gripper fingers. The distance between both contact points consequently has to be smaller than the stroke of the gripper. In addition, the gripper fingers must rest against the outside of the part. Analogue to previous work, the grasps have to be checked for sufficient friction through their angle of contact. Lastly, a simple collision check enables the exclusion of grasp candidates on strongly concave surfaces or near edges. All criteria are depicted from left to right in Fig.~\ref{fig:criteria}.

\subsection{Grasp Synthesis}
\label{sec:grasp_synthesis}

To review these criteria on a given part geometry, we define $g \in \mathrm{SE}(3)$ as a grasp pose, where $G$ is the set of all grasp poses.
Each of these grasp poses is defined by an Euclidean group $\mathrm{SE}(3)$ consisting of a rotation matrix $R\in \mathrm{SO}(3)$ and a translation vector $t\in \mathbb{R}^{3}$. Each point $p$ with translation and normal vector is part of the point cloud $W=\{p_{1},p_{2},\dots,p_{n}\}$. Between any two points $p_{i}$ and $p_{j}$, a grasp pose candidate can be defined. Considering the simplified gripper model to be symmetrical, we can furthermore conclude that $g(p_{i},p_{j})=g(p_{j},p_{i})$. The resulting amount of possible point combinations and therefore grasps leads to $|G|=\frac{1}{2}\cdot(n^{2}-n)$. Accordingly, for a relatively small point cloud of 10,000 points already 49,995,000 point combinations need to be checked. Hence, it is reasonable to examine the criteria in a multi-step procedure. As all criteria need to be fulfilled, point combinations can be excluded if they fail one of the tests.

\begin{figure}
    \centering
    \begin{subfigure}{0.3\textwidth}
        \centering
        \includegraphics[height=1.5cm]{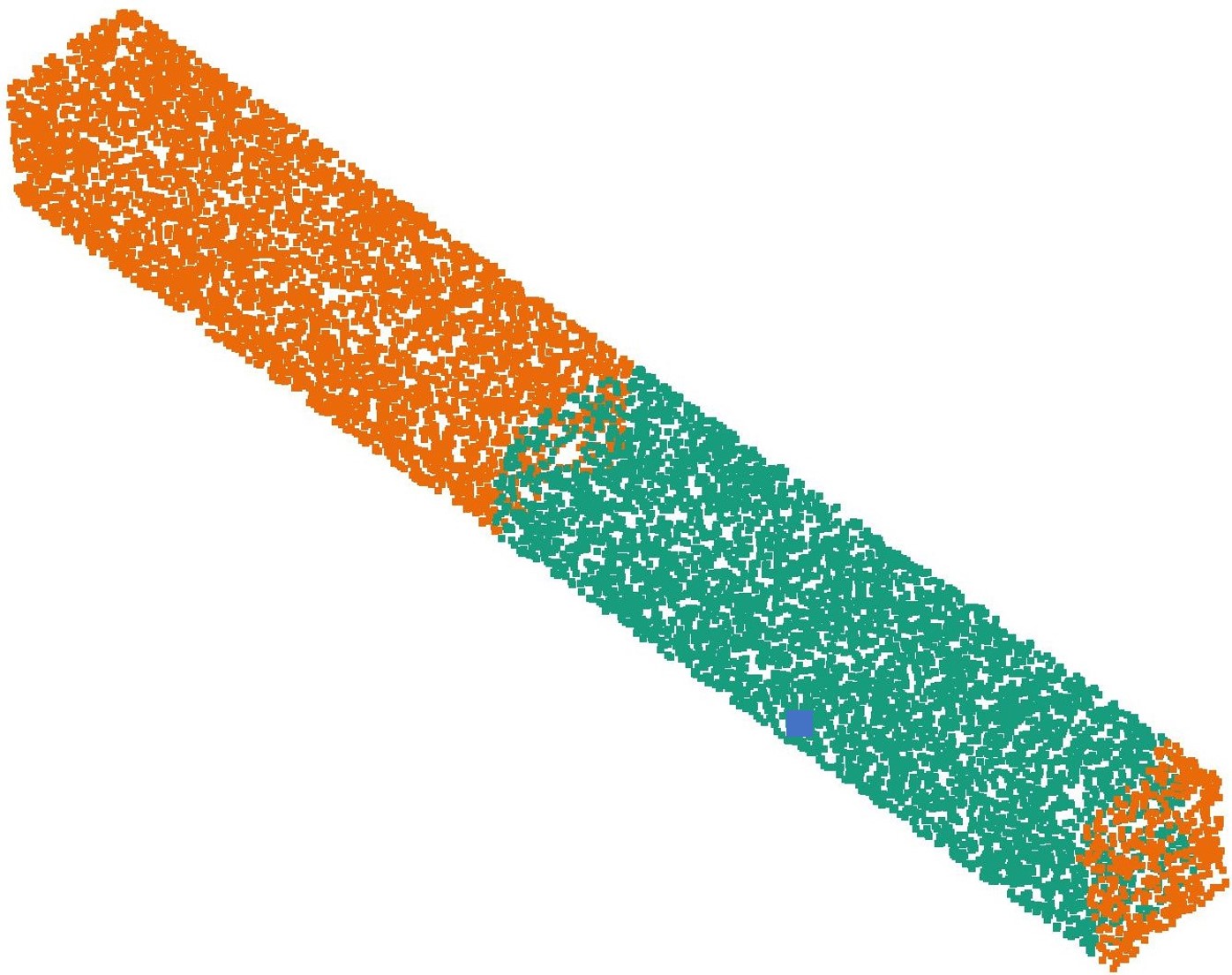}
        \subcaption{}
    \end{subfigure}
    \hfill
    \begin{subfigure}{0.3\textwidth}
        \centering
        \includegraphics[height=1.5cm]{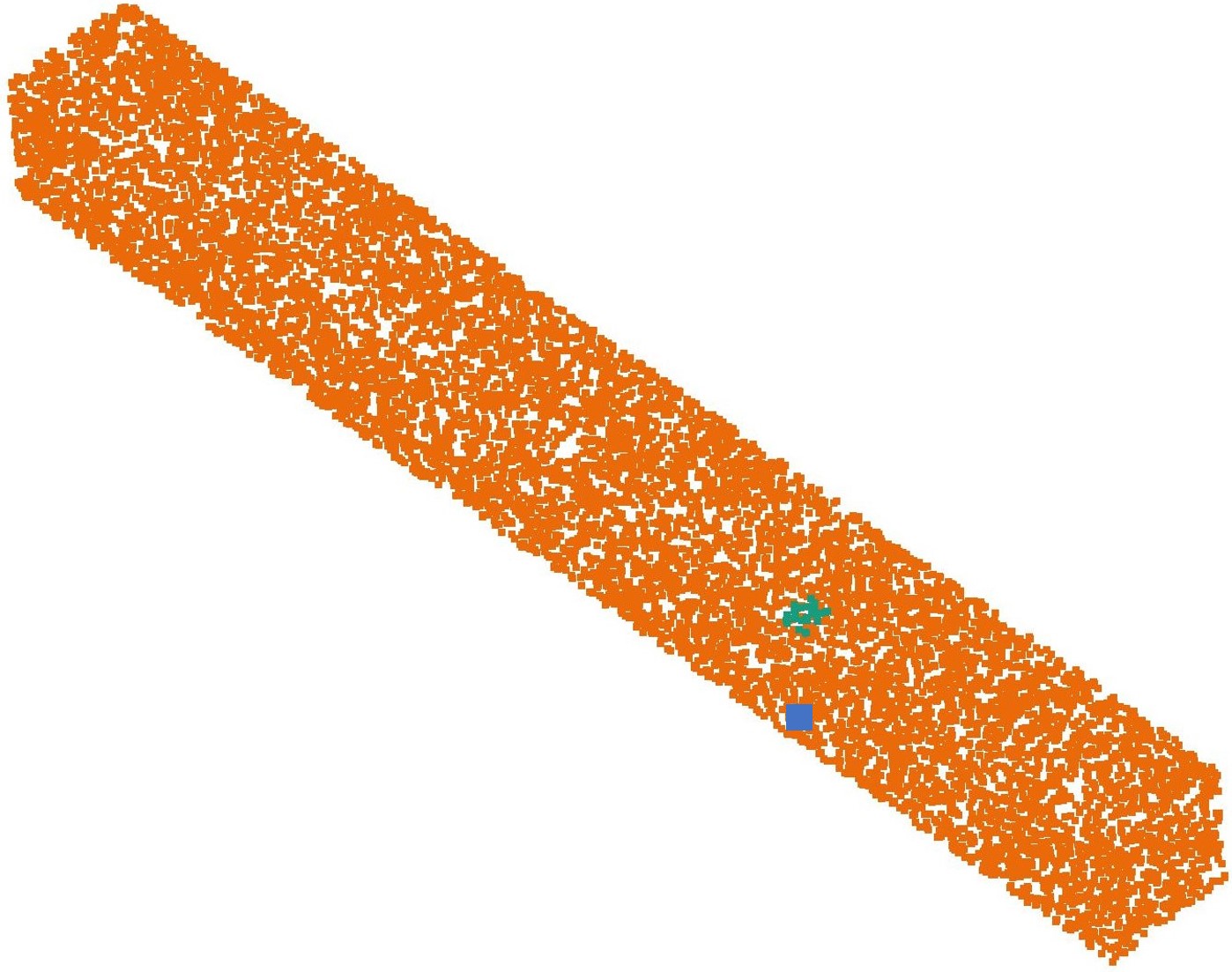}
        \subcaption{}
    \end{subfigure}
    \hfill
    \begin{subfigure}{0.3\textwidth}
        \centering
        \includegraphics[height=1.5cm]{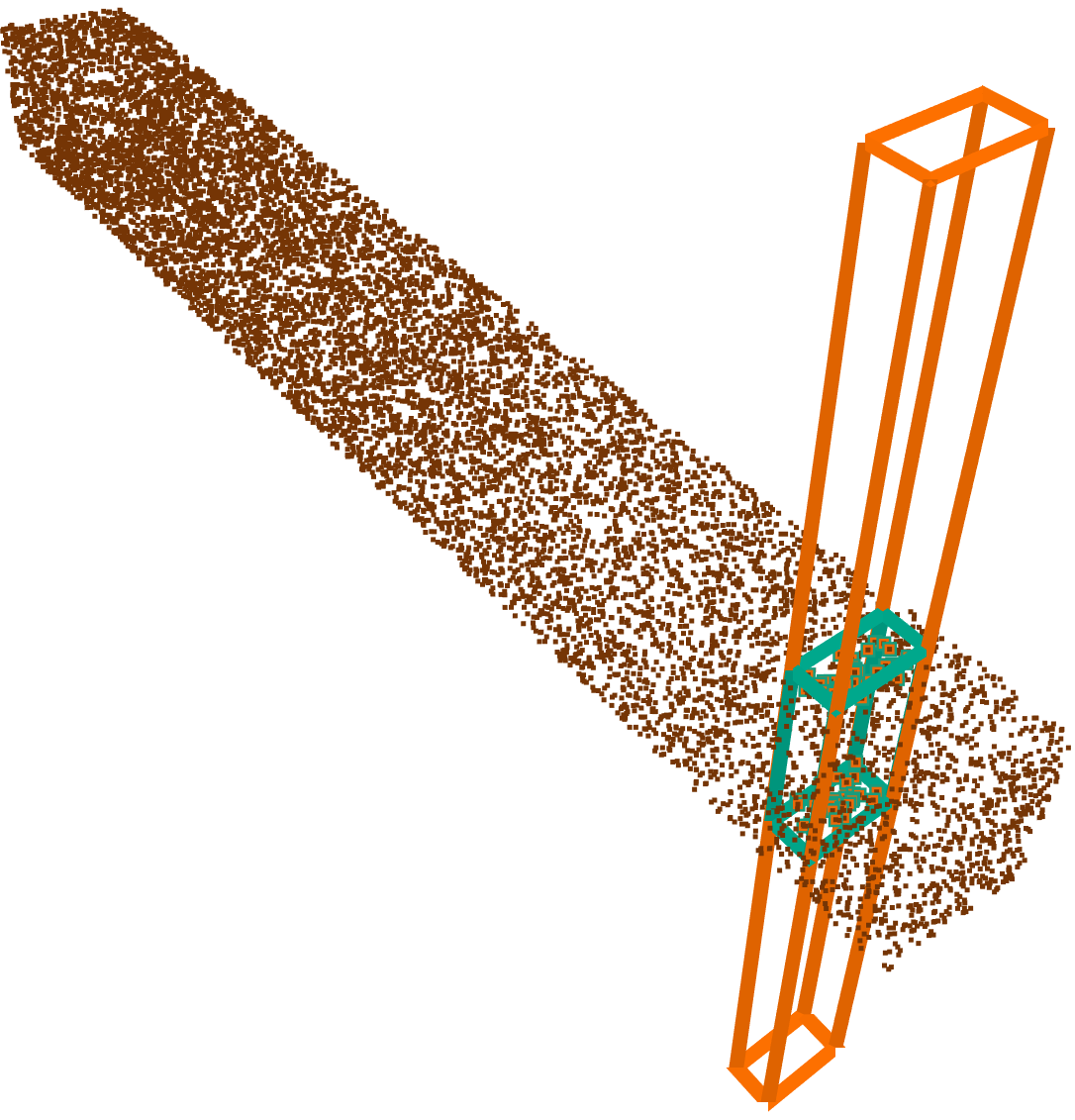}
        \subcaption{}
    \end{subfigure}
    \caption{Conditions visualized on a 3D real point cloud with valid combinations to a blue target point in green and invalid combinations in orange: (a) gripper stroke check, (b) friction check, (c) collision check.}
    \label{fig:criteria_real}
\end{figure}

To check whether the object will fit between the gripper, the distance can be computed as the Euclidean norm of the vector $v \in \mathbb{R}^{3}$ between the translation of the contact points $p$. This is done for every combination of points. Fig.~\ref{fig:criteria_real} shows the results of all possible combinations for a fixed point for the real object named IPABar~\cite{Kleeberger_PQNet.2020} of the \textit{Fraunhofer IPA Bin-Picking dataset}~\cite{Kleeberger_2019,Kleeberger_PQNet.2020}. Point combinations not excluded by the comparison to the stroke of the gripper can then be checked for their contact angles. This is achieved by means of calculating the friction cone of the gripper as shown in Fig.~\ref{fig:contact_model}~(b). The resulting angle of the friction cone must then be compared to the angle between the connecting vector and the surface normal at each contact point. Because we are not yet accounting for the weight, pose, and center of mass of the part, we introduce an additional safety factor of 1.5 to allow for the compensation of acceleration-induced forces such as gravity. Using the connecting vector and surface normals, we can further distinguish grasps from the inside and outside, if needed.

\begin{figure}
    \centering
    \begin{subfigure}{0.3\textwidth}
        \centering
        \includegraphics[height=1.5cm]{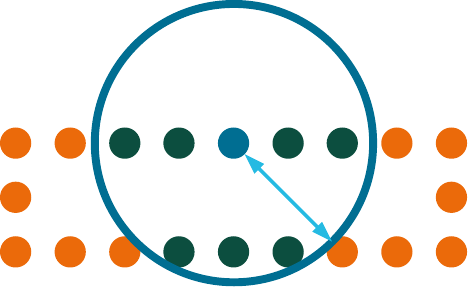}
        \subcaption{}
    \end{subfigure}
    \hfill
    \begin{subfigure}{0.3\textwidth}
        \centering
        \includegraphics[height=1.5cm]{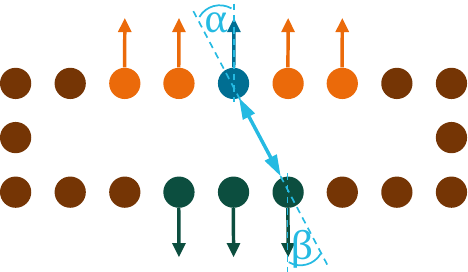}
        \subcaption{}
    \end{subfigure}
    \hfill
    \begin{subfigure}{0.3\textwidth}
        \centering
        \includegraphics[height=1.5cm]{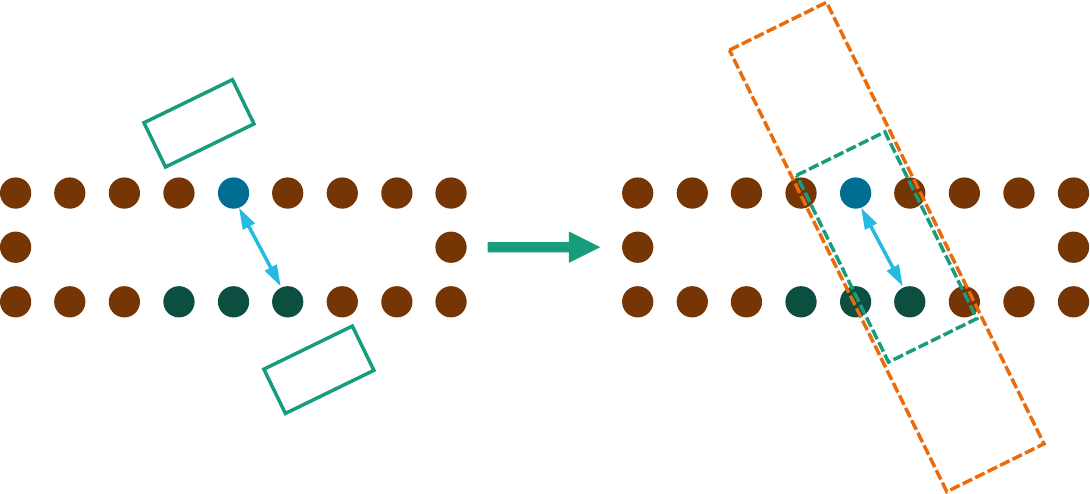}
        \subcaption{}
    \end{subfigure}
    \caption{Conditions visualized on a 2D sample point cloud with valid combinations to a blue target point in green and invalid combinations in orange: (a) gripper stroke check, (b) friction check, (c) collision check.}
    \label{fig:criteria_cloud}
\end{figure}

\begin{figure}[tb]
\centering
\includegraphics[width=0.75\linewidth,keepaspectratio]{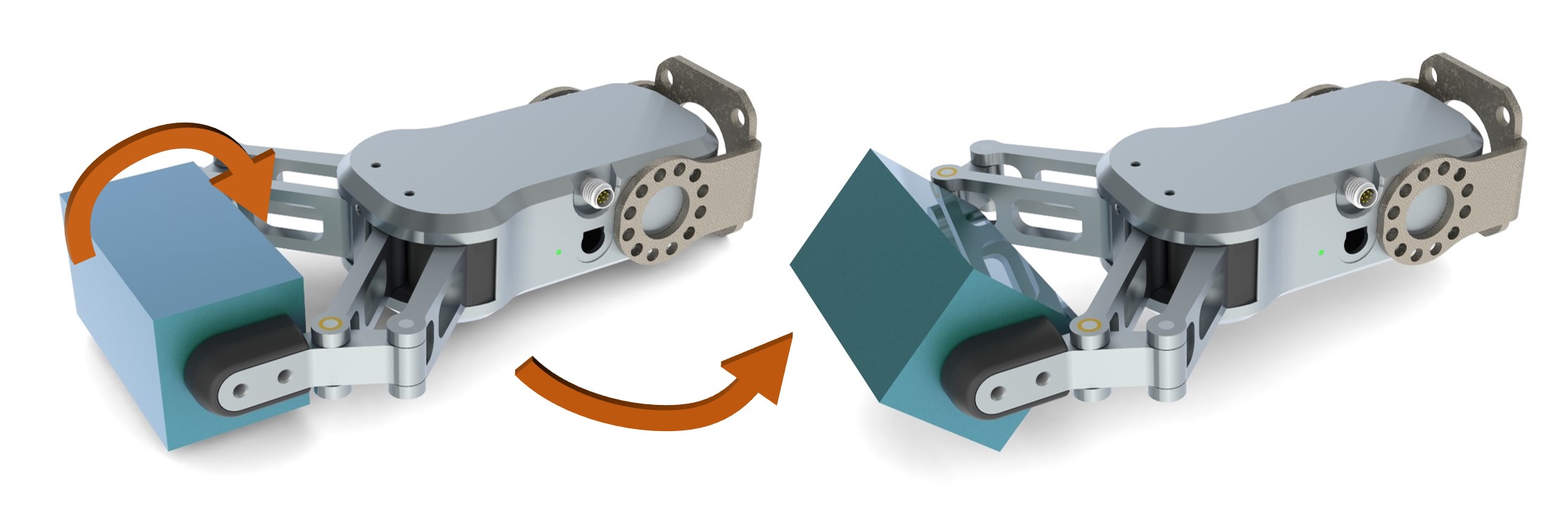}
\caption{Discretization around the middle axis of the gripper.}
\label{fig:discrete_rotation}
\end{figure}

Until now, grasp poses have only been defined by two opposing contact points. With these, we can fully define the grasp pose's translation by simply choosing the midpoint in between. In the same way, two of the three rotational degrees of freedom can be defined. Yet, the rotation around the connecting vector needs to be discretized, as shown in Fig.~\ref{fig:discrete_rotation}. For each discrete orientation, a quick collision check is performed by placing two cuboids
around the contact points. If the set resulting by subtracting the inner (see Fig.~\ref{fig:criteria_cloud}~(c), right or Fig.~\ref{fig:criteria_real}~(c) in orange) from the outer cuboid (see Fig.~\ref{fig:criteria_cloud}~(c), right or Fig.~\ref{fig:criteria_real}~(c) in green) does not contain any points, we can assume that no collision will occur. To account for a small unevenness or curvature of the surface, we set the length of the inner cuboid slightly larger than the norm of the connecting vector. A sketch for each check is shown in Fig.~\ref{fig:criteria_cloud} and a sample image of a real point cloud in Fig.~\ref{fig:criteria_real}.

\subsection{Grasp Clustering}
\label{sec:grasp_clustering}
\begin{figure}
    \centering
    \begin{subfigure}{0.3\textwidth}
        \centering
        \includegraphics[height=1cm]{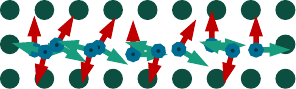}
        \subcaption{}
    \end{subfigure}
    \hfill
    \begin{subfigure}{0.3\textwidth}
        \centering
        \includegraphics[height=1cm]{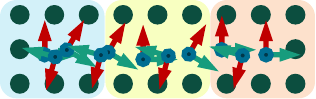}
        \subcaption{}
    \end{subfigure}
    \hfill
    \begin{subfigure}{0.3\textwidth}
        \centering
        \includegraphics[height=1cm]{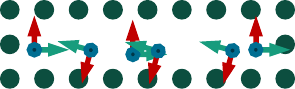}
        \subcaption{}
    \end{subfigure}
    \caption{Grasp reduction through clustering: (a) resulting grasps from sampling, (b) translational clusters, (c) rotational cluster centers and final grasp candidates.}
    \label{fig:clustering}
\end{figure}

\begin{figure}
    \centering
    \begin{subfigure}{0.3\textwidth}
        \centering
        \includegraphics[height=2.5cm]{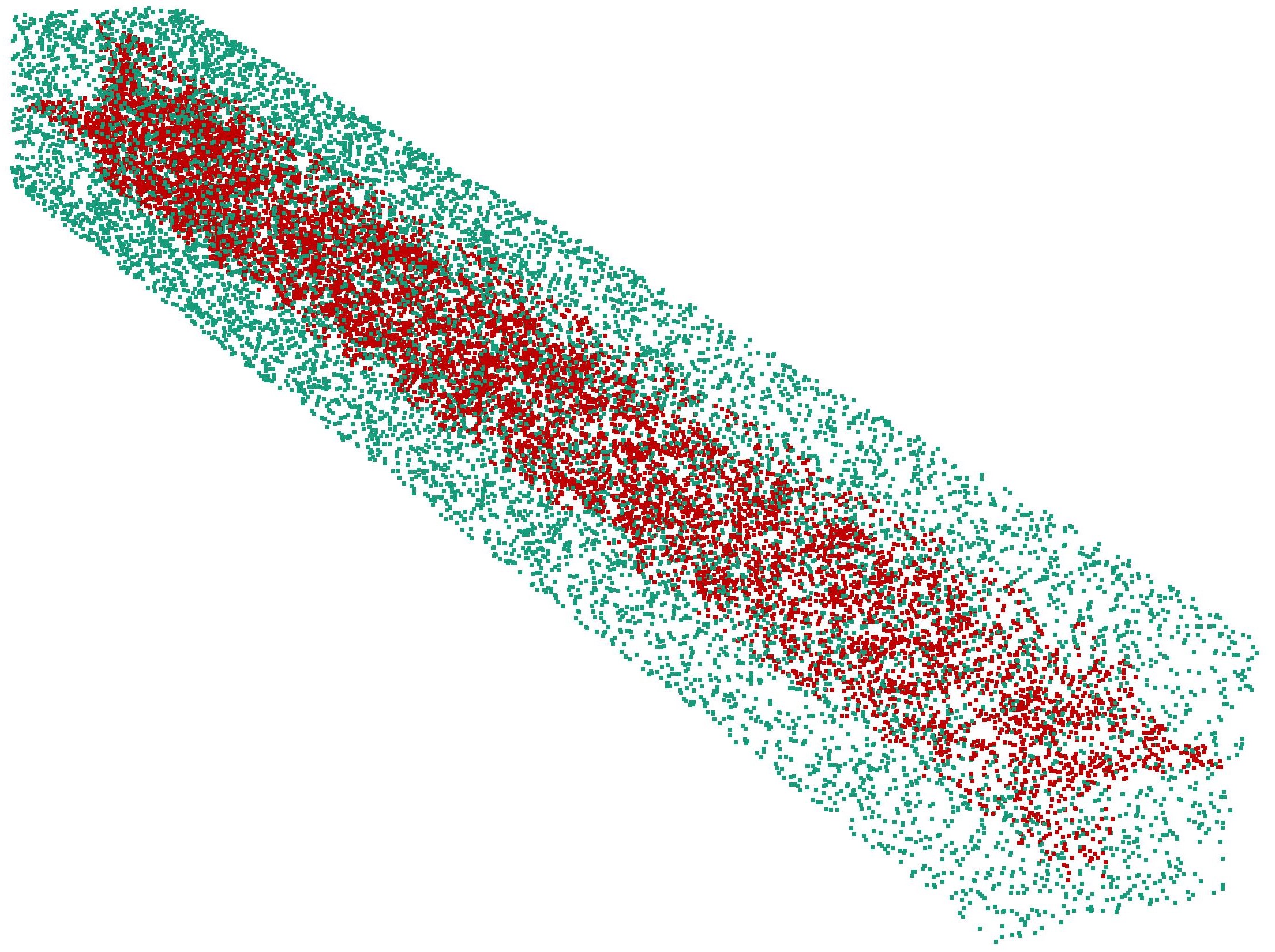}
        \subcaption{}
    \end{subfigure}
    \hfill
    \begin{subfigure}{0.3\textwidth}
        \centering
        \includegraphics[height=2.5cm]{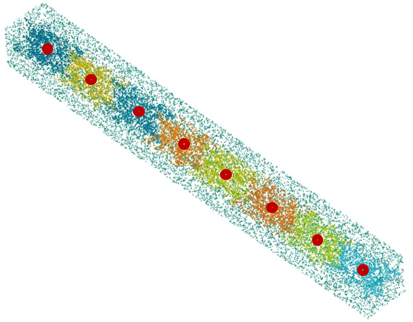}
        \subcaption{}
    \end{subfigure}
    \hfill
    \begin{subfigure}{0.3\textwidth}
        \centering
        \includegraphics[height=2.5cm]{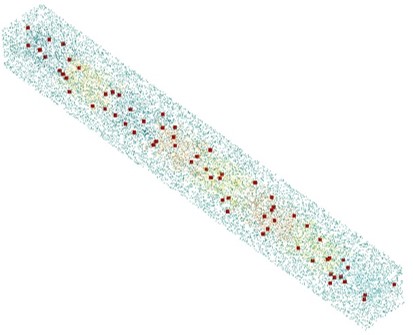}
        \subcaption{}
    \end{subfigure}
    \caption{Grasp reduction through clustering on the IPABar~\cite{Kleeberger_PQNet.2020}: (a) resulting grasps from sampling, (b) translational clusters, (c) rotational cluster centers and final grasp candidates.}
    \label{fig:clustering_real}
\end{figure}

With the described procedure, the amount of possible point combinations can already be significantly reduced. Though, while all of the resulting grasp poses shown in Fig.~\ref{fig:clustering} and Fig.~\ref{fig:clustering_real} might be successful candidates, for most applications several ten thousand grasps are excessive. To further reduce the amount of grasps to be checked in simulation or real-world trials, a clustering algorithm is applied. While many approaches rank the resulting grasps by metrics, such as friction, through this approach we aim to ensure a greater variety of grasps in terms of position and orientation. This optimizes the chance to obtain kinematically feasible grasp poses even in highly cluttered scenes. Solely ranking resulting grasps by some sort of quality metric may result in an accumulation of grasps in some areas, while other areas would remain without any grasps. Reducing the amount of grasps by randomly selecting them could result in a similar problem if there is an accumulation of grasps in a specific area.

Clustering algorithms enable data points to be grouped by their similarity. For our approach, we use a resulting side effect: If clusters contain elements of greater similarity, the dissimilarity between elements of different clusters should be greater. Grasps of the same cluster are nearby in the pose space. Choosing elements of different clusters could accordingly be a feasible approach to ensure a greater diversity of elements. As we are searching for grasp poses in the set of grasps whose selection ensures a great diversity, a $k$-medoids algorithm seems appropriate. In contrast to the more common $k$-means, the cluster centers are actual elements of the clusters instead of their mean. For a first examination, we chose the Partitioning Around Medoids (PAM) algorithm~\cite{Kaufman.1987}. Although rather slow, it proved to be superior to $k$-means for our specific application. Aiming for an offline approach, the slower processing time is negligible.

To ensure a decent distribution of grasps along the geometry of the object, firstly the grasps are clustered according to their midpoint's translation. In a second step, each of the resulting translational clusters is then clustered again according to the normed connecting vector of the point combination. This two-step approach helps ensuring a high variety of graps along the entire part geometry. All described steps are shown schematically in Fig.~\ref{fig:clustering} and on the sample object IPABar~\cite{Kleeberger_PQNet.2020} in Fig.~\ref{fig:clustering_real}.

\begin{figure}[tb]
    \centering
    \begin{subfigure}{0.2\textwidth}
        \centering
        \includegraphics[width=\textwidth]{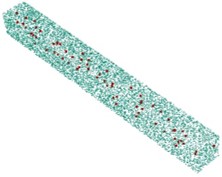}
        \subcaption{}
    \end{subfigure}
    \hfill
    \begin{subfigure}{0.2\textwidth}
        \centering
        \includegraphics[width=\textwidth]{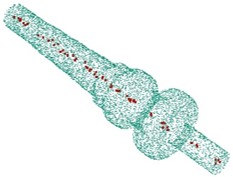}
        \subcaption{}
    \end{subfigure}
    \hfill
    \begin{subfigure}{0.2\textwidth}
        \centering
        \includegraphics[width=\textwidth]{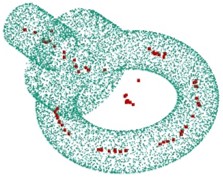}
        \subcaption{}
    \end{subfigure}
    \hfill
    \begin{subfigure}{0.15\textwidth}
        \centering
        \includegraphics[width=\textwidth]{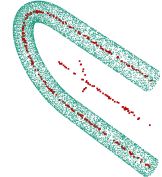}
        \subcaption{}
    \end{subfigure}
    
    \medskip
    
    \centering
    \begin{subfigure}{0.2\textwidth}
        \centering
        \includegraphics[width=\textwidth]{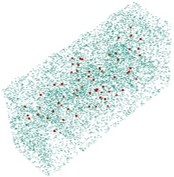}
        \subcaption{}
    \end{subfigure}
    \hfill
    \begin{subfigure}{0.2\textwidth}
        \centering
        \includegraphics[width=\textwidth]{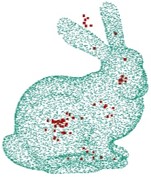}
        \subcaption{}
    \end{subfigure}
    \hfill
    \begin{subfigure}{0.2\textwidth}
        \centering
        \includegraphics[width=\textwidth]{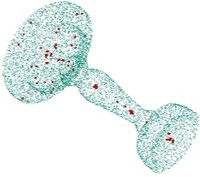}
        \subcaption{}
    \end{subfigure}
    \hfill
    \begin{subfigure}{0.2\textwidth}
        \centering
        \includegraphics[width=\textwidth]{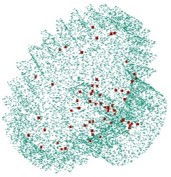}
        \subcaption{}
    \end{subfigure}
    
    \medskip
    
    \centering
    \begin{subfigure}{0.2\textwidth}
        \centering
        \includegraphics[width=\textwidth]{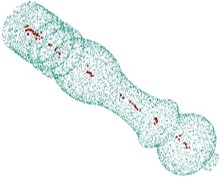}
        \subcaption{}
    \end{subfigure}
    \hfill
    \begin{subfigure}{0.2\textwidth}
        \centering
        \includegraphics[width=\textwidth]{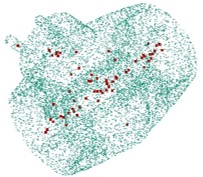}
        \subcaption{}
    \end{subfigure}
    \hfill
    \begin{subfigure}{0.2\textwidth}
        \centering
        \includegraphics[width=\textwidth]{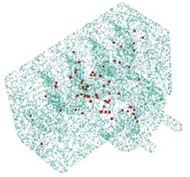}
        \subcaption{}
    \end{subfigure}
    \hfill
    \begin{subfigure}{0.2\textwidth}
        \centering
        \includegraphics[width=\textwidth]{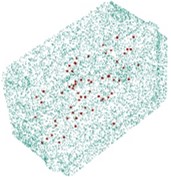}
        \subcaption{}
    \end{subfigure}
    
    \hfill
    
    \caption{Example geometries of the Fraunhofer IPA~\cite{Kleeberger_2019} and Sil\'{e}ane~\cite{Bregier.2017} datasets. The point sampled geometry is shown as green dots and the resulting grasp
    points are displayed in red for the
    (a)~\mbox{IPABar}, (b)~\mbox{IPAGearShaft}, (c)~\mbox{IPARingScrew}, (d)~\mbox{IPAUBolt}, (e)~\mbox{SileaneBrick}, (f)~\mbox{SileaneBunny}, (g)~\mbox{SileaneCandleStick}, (h)~\mbox{SileaneGear}, (i)~\mbox{SileanePepper}, (j)~\mbox{SileaneTLess20}, (k)~\mbox{SileaneTLess22}, and (l)~\mbox{SileaneTLess29} object.}
    \label{fig:example_geometries}
\end{figure}

\section{Evaluation}
\label{sec:evaluation}

\subsection{Qualitative Analysis}
Applying the described algorithm to various parts of the \textit{Fraunhofer IPA Bin-Picking dataset}~\cite{Kleeberger_2019} and the \textit{Sil\'{e}ane Dataset for Object Detection and Pose Estimation}~\cite{Bregier.2017} yielded the results shown in Fig.~\ref{fig:example_geometries}. A solely qualitative analysis of the synthesized grasps supports the validity of our proposed approach. Grasp pose candidates could be found for every tested object and the resulting grasps are spread
all over the geometry of the part.

\subsection{Experiments in a Real-World Robot Cell}
\label{sec:experiments}






To evaluate our approach in a real-world task, an analytical bin-picking solution\footnote{\url{https://www.youtube.com/watch?v=xhTkgajg8wQ}}~\cite{Spenrath.2013} was provided with the option to import the generated grasps.
The approach records a 3D point cloud of the scene, estimates 6D object poses using template matching~\cite{Diss_Ledermann.2012,Diss_Palzkill.2014}, and selects a collision-free and kinematically feasible grasp pose on the localized objects using heuristic search~\cite{FPS_GPC_1,FPS_GPC_2,Spenrath.2017}.

Previously, possible grasp poses had to be manually defined on the object of choice. Using the presented approach, this time-consuming step can be neglected. The grasping trials show a successful application of different grasp poses generated by the proposed algorithm. Using the grasps generated by our approach, we report a grasp success rate of 91\% during the emptying process of 25 chaotically filled bins with ring screws (see Fig. \ref{fig:example_geometries}~(c)). Videos of our experiments are available at \url{https://owncloud.fraunhofer.de/index.php/s/MPpjLHgubg5f2dV}.

\subsection{Discussion}
\label{sec:discussion}
In the following, we summarize strengths and discuss limitations of our approach.

\subsubsection{Strengths}
The use of point cloud data as input format enables us not only to be independent of the various existing CAD formats, but also allows for the immediate use of part scans if no CAD data is available. In industrial usage, we can expect to either obtain CAD data or a part scan beforehand. For reasonable point cloud sizes, we can generate a great amount of grasp candidates in a matter of seconds to few minutes on a common office computer. Further improvements in computation time are possible through the utilization of task parallelization. 

A first, qualitative analysis of the generated grasps on different parts shows a good distribution over the geometry of the object. Through the use of the PAM clustering algorithm, we can also obtain a high variance in grasp poses. To our knowledge, this is the first implementation of a clustering algorithm in search of a reduced set of highly diverse grasps. By simplifying the gripper's main geometric parameters, our algorithm is further fairly easy to use and adapt to different grippers. The application in
a real-world robot cell proved to be promising. For the generation of grasp poses, almost no user interaction is required, resulting in a highly efficient generation of grasp candidates.

\subsubsection{Limitations}
Whereas the time needed for the generation of grasp poses is reasonable for an offline approach, in comparison to online approaches it is still rather slow. Although enabling a great variance in grasp poses, the clustering algorithm takes up a major amount of the time needed for grasp generation. The computation time of the clustering algorithm rises with an increasing amount of grasp poses. In our tests on a common office computer, the computation time for the clustering ranged from a few minutes up to a couple of hours. To accelerate the process, multiple approaches are feasible. Firstly, the clustering algorithm is a great candidate for task parallelization, as is the grasp generation algorithm. Further, another, less precise clustering algorithm could be used. First trials with $k$-Means~\cite{Lloyd.1982} showed limited success, as clustering seemed sub-par compared to the results acquired with $k$-Medoids. A reduction of grasps fed into the clustering algorithm is another viable way, although problems were mentioned in Section~\ref{sec:auto_generation} when scaling down the cloud by a randomly applied filter.

The collision check implemented is rather minimal. Using a full gripper model would allow for the algorithm to reduce grasp poses more efficiently. The full integration of a gripper model is planned for the integration of the algorithm in a holistic bin-picking approach.


Our method allows to report high grasping success rates in real-world robot systems. Still, the approach does not consider a physical interaction between the object and the gripper. Testing the grasps in a physics simulation or removing unsuccessful grasps on the real system from the set of possible grasps are options to further increase the grasping success rate.

\section{Conclusions and Future Work}
\label{sec:conclusion}
In this paper, we proposed a novel approach for the automatic generation of grasp poses for known rigid objects for parallel jaw grippers.
Using point clouds of the objects to be grasped, various metrics are checked to find possible grasp candidates.
To further reduce the grasp poses located by this process, we use a clustering algorithm to refine the created set of grasps while maintaining a high variance in grasp poses.
In future work,
the approach shall be extended to other gripper types, such as
suction
and magnetic grippers.

\section*{Acknowledgment}
This work was partially supported by the Federal Ministry of Education and Research (Deep Picking -- Grant No. 01IS20005C) and the State Ministry of Baden-W\"urttemberg for Economic Affairs, Labour and Housing Construction (Center for Cognitive Robotics –- Grant No. 017-180004 and Center for Cyber Cognitive Intelligence (CCI) -- Grant No. 017-192996).


\bibliographystyle{IEEEtran}
\bibliography{references}

\begin{thebibliography}{10}
\providecommand{\url}[1]{#1}
\csname url@samestyle\endcsname
\providecommand{\newblock}{\relax}
\providecommand{\bibinfo}[2]{#2}
\providecommand{\BIBentrySTDinterwordspacing}{\spaceskip=0pt\relax}
\providecommand{\BIBentryALTinterwordstretchfactor}{4}
\providecommand{\BIBentryALTinterwordspacing}{\spaceskip=\fontdimen2\font plus
\BIBentryALTinterwordstretchfactor\fontdimen3\font minus
  \fontdimen4\font\relax}
\providecommand{\BIBforeignlanguage}[2]{{%
\expandafter\ifx\csname l@#1\endcsname\relax
\typeout{** WARNING: IEEEtran.bst: No hyphenation pattern has been}%
\typeout{** loaded for the language `#1'. Using the pattern for}%
\typeout{** the default language instead.}%
\else
\language=\csname l@#1\endcsname
\fi
#2}}
\providecommand{\BIBdecl}{\relax}
\BIBdecl

\bibitem{Litvak.2019}
Y.~Litvak, A.~Biess, and A.~Bar-Hillel, ``Learning pose estimation for
  high-precision robotic assembly using simulated depth images,'' in \emph{IEEE
  International Conference on Robotics and Automation (ICRA)}, 2019.

\bibitem{Wolf.2005}
A.~Wolf, R.~Steinmann, and H.~Schunk, \emph{Grippers in motion: The fascination
  of automated handling tasks}.\hskip 1em plus 0.5em minus 0.4em\relax Berlin:
  Springer, 2005.

\bibitem{MolzowVoit.2016}
F.~Molzow-Voit, M.~Quandt, M.~Freitag, and G.~Sp{\"o}ttl, \emph{Robotik in der
  Logistik: Qualifizierung f{\"u}r Fachkr{\"a}fte und Entscheider},
  1st~ed.\hskip 1em plus 0.5em minus 0.4em\relax Wiesbaden: Springer, 2016.

\bibitem{Muller.2019}
R.~M{\"u}ller, J.~Franke, D.~Henrich, B.~Kuhlenk{\"o}tter, A.~Raatz, and
  A.~Verl, \emph{Handbuch Mensch-Roboter-Kollaboration}.\hskip 1em plus 0.5em
  minus 0.4em\relax M{\"u}nchen: Hanser, 2019.

\bibitem{VogelHeuser.2017}
B.~Vogel-Heuser, T.~Bauernhansl, and M.~ten Hompel, \emph{Handbuch Industrie
  4.0 Bd.2: Automatisierung}, 2nd~ed., ser. Springer.\hskip 1em plus 0.5em
  minus 0.4em\relax Berlin: Springer, 2017.

\bibitem{Buchholz.2016}
D.~Buchholz, \emph{Bin-Picking: New Approaches for a Classical Problem},
  1st~ed., ser. Studies in Systems, Decision and Control.\hskip 1em plus 0.5em
  minus 0.4em\relax Cham: Springer, 2016, vol.~44.

\bibitem{Kleeberger_PQNet.2020}
K.~Kleeberger, M.~V{\"o}lk, M.~Moosmann, E.~Thiessenhusen, F.~Roth, R.~Bormann,
  and M.~F. Huber, ``Transferring experience from simulation to the real world
  for precise pick-and-place tasks in highly cluttered scenes,'' in
  \emph{IEEE/RSJ International Conference on Intelligent Robots and Systems
  (IROS)}, 2020.

\bibitem{Kleeberger_ReviewArticle.2020}
K.~Kleeberger, R.~Bormann, W.~Kraus, and M.~F. Huber, ``A survey on
  learning-based robotic grasping,'' in \emph{Current Robotics Reports}, J.~N.
  Pires, Ed.\hskip 1em plus 0.5em minus 0.4em\relax Springer, 2020.

\bibitem{Pas.2015}
A.~t. Pas and R.~Platt, ``Using geometry to detect grasps in 3d point clouds,''
  \emph{International Symposium on Robotics Research (ISRR)}, 2015.

\bibitem{Shao.2018}
Q.~Shao and J.~Hu, ``Combining rgb and points to predict grasping region for
  robotic bin-picking,'' \emph{arXiv:1904.07394}, 2018.

\bibitem{ZapataImpata.2019}
B.~S. Zapata-Impata, P.~Gil, J.~Pomares, and F.~Torres, ``Fast geometry-based
  computation of grasping points on three-dimensional point clouds,''
  \emph{International Journal of Advanced Robotic Systems}, vol.~16, no.~1,
  2019.

\bibitem{Kundu.2018}
O.~Kundu and S.~Kumar, ``Optimized edge-based grasping method for a cluttered
  environment,'' \emph{arXiv:1809.01243}, 2018.

\bibitem{MA_FlorianRoth}
F.~Roth, ``Development of a model-based heuristic for automated grasp pose
  determination with the help of machine learning,'' master's thesis,
  {University of Stuttgart}, Stuttgart, 2020.

\bibitem{Borst.2003}
C.~Borst, M.~Fischer, and G.~Hirzinger, ``Grasping the dice by dicing the
  grasp,'' in \emph{IEEE/RSJ International Conference on Intelligent Robots and
  Systems (IROS)}, 2003.

\bibitem{Siciliano.2016}
B.~Siciliano and O.~Khatib, \emph{Springer Handbook of Robotics}, 2nd~ed.\hskip
  1em plus 0.5em minus 0.4em\relax Berlin: Springer, 2016.

\bibitem{Correll.2016}
N.~Correll, K.~E. Bekris, D.~Berenson, O.~Brock, A.~Causo, K.~Hauser, K.~Okada,
  A.~Rodriguez, J.~M. Romano, and P.~R. Wurman, ``Analysis and observations
  from the first amazon picking challenge,'' \emph{IEEE Transactions on
  Automation Science and Engineering (TASE)}, vol.~15, no.~1, pp. 172--188,
  2016.

\bibitem{Sahbani.2012}
A.~Sahbani, S.~El-Khoury, and P.~Bidaud, ``An overview of 3d object grasp
  synthesis algorithms,'' \emph{Robotics and Autonomous Systems}, vol.~60,
  no.~3, pp. 326--336, 2012.

\bibitem{Bohg.2014}
J.~Bohg, A.~Morales, T.~Asfour, and D.~Kragic, ``Data-driven grasp
  synthesis---a survey,'' \emph{IEEE Transactions on Robotics (T-RO)}, vol.~30,
  no.~2, pp. 289--309, 2014.

\bibitem{Johns.2016}
E.~Johns, S.~Leutenegger, and A.~J. Davison, ``Deep learning a grasp function
  for grasping under gripper pose uncertainty,'' in \emph{IEEE/RSJ
  International Conference on Intelligent Robots and Systems (IROS)}, 2016.

\bibitem{Mahler.2016}
J.~Mahler, F.~T. Pokorny, B.~Hou, M.~Roderick, M.~Laskey, M.~Aubry,
  K.~Kohlhoff, T.~Kr{\"o}ger, J.~Kuffner, and K.~Goldberg, ``Dex-net 1.0: A
  cloud-based network of 3d objects for robust grasp planning using a
  multi-armed bandit model with correlated rewards,'' in \emph{IEEE
  International Conference on Robotics and Automation (ICRA)}, 2016.

\bibitem{Mousavian.2019}
A.~Mousavian, C.~Eppner, and D.~Fox, ``6-dof graspnet: Variational grasp
  generation for object manipulation,'' in \emph{IEEE International Conference
  on Computer Vision (ICCV)}, 2019.

\bibitem{Bohg.2011}
J.~Bohg, M.~Johnson-Roberson, B.~Le{\'o}n, J.~Felip, X.~Gratal,
  N.~Bergstr{\"o}m, D.~Kragic, and A.~Morales, ``Mind the gap - robotic
  grasping under incomplete observation,'' in \emph{IEEE International
  Conference on Robotics and Automation (ICRA)}, 2011.

\bibitem{Brook.2011}
P.~Brook, M.~Ciocarlie, and K.~Hsiao, ``Collaborative grasp planning with
  multiple object representations,'' in \emph{IEEE International Conference on
  Robotics and Automation (ICRA)}, 2011.

\bibitem{FPS_GPC_1}
F.~Spenrath, A.~Spiller, and A.~Verl, ``Gripping point determination and
  collision prevention in a bin-picking application,'' in \emph{German
  Conference on Robotics (ROBOTIK)}, 2012.

\bibitem{FPS_GPC_2}
F.~Spenrath and A.~Pott, ``Gripping point determination for bin picking using
  heuristic search,'' in \emph{CIRP Conference on Intelligent Computation in
  Manufacturing Engineering (CIRP ICME)}, 2016.

\bibitem{Spenrath.2017}
------, ``Statistical analysis of influencing factors for heuristic grip
  determination in random bin picking,'' in \emph{IEEE International Conference
  on Advanced Intelligent Mechatronics (AIM)}, 2017.

\bibitem{Buchholz.2013}
D.~Buchholz, M.~Futterlieb, S.~Winkelbach, and F.~M. Wahl, ``Efficient
  bin-picking and grasp planning based on depth data,'' in \emph{IEEE
  International Conference on Robotics and Automation (ICRA)}, 2013.

\bibitem{Spenrath.2013}
F.~Spenrath, M.~Palzkill, A.~Pott, and A.~Verl, ``Object recognition:
  Bin-picking for industrial use,'' in \emph{IEEE International Symposium on
  Robotics (ISR)}, 2013.

\bibitem{Chen.2011}
C.-H. Chen, H.-P. Huang, and {Lo Sheng-Yen}, ``Stereo-based 3d localization for
  grasping known objects with a robotic arm system,'' \emph{World Congress on
  Intelligent Control and Automation (WCICA)}, pp. 309--314, 2011.

\bibitem{Schraft.2003}
R.~D. Schraft and T.~Ledermann, ``Intelligent picking of chaotically stored
  objects,'' \emph{Assembly Automation}, vol.~23, no.~1, pp. 38--42, 2003.

\bibitem{Mahler.2015}
J.~Mahler, S.~Patil, B.~Kehoe, J.~{van den Berg}, M.~Ciocarlie, P.~Abbeel, and
  K.~Goldberg, ``Gp-gpis-opt: Grasp planning with shape uncertainty using
  gaussian process implicit surfaces and sequential convex programming,'' in
  \emph{IEEE International Conference on Robotics and Automation (ICRA)}, 2015.

\bibitem{Gatrell.1989}
L.~B. Gatrell, ``Cad-based grasp synthesis utilizing polygons, edges and
  vertexes,'' in \emph{IEEE International Conference on Robotics and Automation
  (ICRA)}, 1989.

\bibitem{Smith.1999}
G.~Smith, E.~Lee, K.~Goldberg, K.~B{\"o}hringer, and J.~Craig, ``Computing
  parallel-jaw grips,'' in \emph{IEEE International Conference on Robotics and
  Automation (ICRA)}, 1999.

\bibitem{Depierre.2018}
A.~Depierre, E.~Dellandr{\'e}a, and L.~Chen, ``Jacquard: A large scale dataset
  for robotic grasp detection,'' in \emph{IEEE/RSJ International Conference on
  Intelligent Robots and Systems (IROS)}, 2018.

\bibitem{Saxena.2008}
A.~Saxena, J.~Driemeyer, and A.~Y. Ng, ``Robotic grasping of novel objects
  using vision,'' \emph{The International Journal of Robotics Research (IJRR)},
  vol.~27, no.~2, pp. 157--173, 2008.

\bibitem{Ikeuchi.1983}
K.~Ikeuchi, B.~K. Horn, S.~Nagata, T.~Callahan, and O.~Feingold, ``Picking up
  an object from a pile of objects,'' \emph{Proceedings of the First
  International Symposium on Robotics Research}, 1983.

\bibitem{Horn.1984}
B.~Horn, ``Extended gaussian images,'' \emph{Proceedings of the IEEE}, vol.~72,
  no.~12, pp. 1671--1686, 1984.

\bibitem{Hirai.1991}
S.~Hirai, ``Analysis and planning of manipulation using the theory of
  polyhedral convex cones,'' Dissertation, {Kyoto University}, Kyoto, 1991.

\bibitem{Bormann.2019}
R.~Bormann, B.~F. de~Brito, J.~Lindermayr, M.~Omainska, and M.~Patel, ``Towards
  automated order picking robots for warehouses and retail,'' in
  \emph{International Conference on Computer Vision Systems (ICVS)}, 2019.

\bibitem{Kasper.2012}
A.~Kasper, Z.~Xue, and R.~Dillmann, ``The kit object models database: An object
  model database for object recognition, localization and manipulation in
  service robotics,'' \emph{The International Journal of Robotics Research
  (IJRR)}, vol.~31, no.~8, pp. 927--934, 2012.

\bibitem{Kleeberger_2019}
K.~Kleeberger, C.~Landgraf, and M.~F. Huber, ``Large-scale 6d object pose
  estimation dataset for industrial bin-picking,'' in \emph{IEEE/RSJ
  International Conference on Intelligent Robots and Systems (IROS)}, 2019.

\bibitem{Kaufman.1987}
L.~Kaufman and P.~J. Rousseeuw, ``Clustering by means of medoids,''
  \emph{Statistical Data Analysis Based on the L1--Norm and Related Methods},
  1987.

\bibitem{Bregier.2017}
\BIBentryALTinterwordspacing
R.~Bregier, F.~Devernay, L.~Leyrit, and J.~L. Crowley, ``Sil\'{e}ane dataset
  for object detection and pose estimation,'' 2017. [Online]. Available:
  \url{https://rbregier.github.io/dataset2017}
\BIBentrySTDinterwordspacing

\bibitem{Diss_Ledermann.2012}
T.~Ledermann, ``{Partikel-Schwarm-Optimierung zur Objektlageerkennung in
  Tiefendaten},'' Dissertation, {University of Stuttgart}, Stuttgart, 2012.

\bibitem{Diss_Palzkill.2014}
M.~Palzkill, ``{Heuristisches Suchverfahren zur Objektlageerkennung aus
  Punktewolken f{\"u}r industrielle Zuf{\"u}hrsysteme},'' Dissertation,
  {University of Stuttgart}, Stuttgart, 2014.

\bibitem{Lloyd.1982}
S.~P. Lloyd, ``Least squares quantization in pcm,'' \emph{IEEE Transactions on
  Information Theory (TIT)}, vol.~28, no.~2, pp. 129--137, 1982.

\end{thebibliography}

\end{document}